# Deep Learning Techniques for Phoneme Recognition in Italian Children's Speech

Nicola Barbaro[a], Cristina Gena[a], Francesco Petriglia[b], Andrea Meirone[b], Alessandro Mazzei[a], Arianna Viotti[b]

*a University of Turin, Computer Science Department, Via Pessinetto 12, 10149 Turin, Italy*

*b Fondazione Paideia Ente Filantropico, Via Moncalvo 1, 10131 Turin, Italy*

ORCID: 0009-0003-8467-2946 (N. Barbaro); 0000-0003-0049-6213 (C. Gena); 0000-0003-0459-8225 (F. Petriglia); 0009-0007-1265-1942 (A. Meirone); 0000-0003-3072-0108 (A. Mazzei); 0009-0005-6637-8377 (A. Viotti)

## Abstract

Speech therapists often face difficulties diagnosing impairments due to the lack of efficient tools for transcribing speech into the International Phonetic Alphabet (IPA). This work addresses this challenge with Broca, a Conformer-based deep learning system pretrained on 8 days of adult speech and fine-tuned on a 165-minute dataset of Italian child speech collected through a range of standardized diagnostic tests for children aged 3.5–6.5. Broca was optimized to handle phonetic variability in children's speech, including tone, accent, and speech errors, and achieved a state-of-the-art weighted Phoneme Error Rate of 13.36% on Italian speech. Remarkably, this performance was obtained using less than three hours of child-specific data, underscoring the model's efficiency and robustness in low-resource clinical settings. This work demonstrates that accurate, vocabulary-independent speech-to-IPA transcription can be achieved with minimal data, paving the way for more accessible, data-efficient tools to support speech assessment and diagnosis.

## 1. Introduction

Speech therapy is a specialized field dedicated to assessing and treating communication disorders, speech impairments, and language difficulties across individuals of all ages. Communication skills play a central role in early cognitive, social, and academic development. Disorders in speech and language can significantly affect a child's ability to express thoughts, understand others, and integrate socially and academically. Research has shown that children with developmental Language Disorders (LD) are at higher risk for social, emotional, and academic challenges. For example, a study by Harel et al. [35] found that children with residual speech errors (i.e. speech errors that continue past the age of 9) experienced more difficulties compared to their peers with typical speech development. Similarly, Gertner et al. found in [29] that children with typical language development language skills were more likely to be positively received by their peers, while those with LD or English as a second language backgrounds were often categorized as "Disliked" or "Low Impact". Vocabulary skills, in particular, strongly predicted peer popularity, emphasizing how crucial communication competence is for children's social acceptance.

Speech-Language Pathologists (SLPs) play a central role in the assessment and treatment of communication disorders. Their intervention focuses on impairments in language comprehension and production, including difficulties in understanding spoken language, constructing grammatically correct sentences, organizing discourse, and using language appropriately in social contexts. They also address deficits in speech sound development which may affect speech intelligibility and overall communicative effectiveness. These challenges may occur in developmental conditions such as autism spectrum disorders (ASD) and LD, as well as in acquired or neurogenic conditions. Early intervention by SLPs has been proven to significantly improve children's language comprehension, production skills, and social integration [30]. Accurate assessment is essential for effective therapy.

As Betz highlighted in [8], clinicians must carefully select standardized tests that are reliable, valid, and sensitive. They must also consider cultural and linguistic backgrounds, time constraints, and resource availability. Cultural diversity plays a critical role in therapy: linguistic differences across cultures can influence both the manifestation of speech disorders and the strategies used to treat them. Verdon [63] emphasized the necessity for SLPs to develop cultural competence to ensure equitable and effective therapy outcomes.

Language structure also impacts how impairments manifest. Leonard [42] noted that symptoms of LD differ across languages depending on their grammatical and phonological characteristics. Moreover, the regional variants of a language can complicate the recognition and treatment of LD. For example, Italian dicts, which often modify or omit grammatical elements like clitics, further complicate assessment and diagnosis. Indeed, Bortolini et al. [12] identified direct-object clitics and non-word repetitions as strong markers for identifying LD in Italian-speaking children, but dictal variation must be carefully distinguished from true impairment.

One of the most widely recommended and ecologically valid methods for assessing expressive language in children is language sample analysis (LSA). This approach involves the systematic collection and analysis of spontaneous language produced by the child in naturalistic or semi-structured contexts. LSA provides detailed and clinically relevant information across multiple linguistic domains, including morphosyntactic structures, lexical diversity, utterance length, and discourse organization. Unlike many standardized tests, LSA captures functional language use and allows for individualized, developmentally sensitive profiling of language abilities. However, despite its recognized clinical utility, LSA is underutilized in many school-based and clinical settings. As reported by Lenhart et al. in [41], practitioners frequently cite time constraints, lack of resources, and insufficient training as major barriers to its implementation. The process of transcribing, coding, and analyzing language samples is often perceived as labor-intensive and technically demanding, which limits its routine adoption, particularly in contexts with high caseloads and limited support infrastructure. Addressing these barriers remains essential for integrating LSA more fully into evidence-based practice. An additional factor contributing to the complexity of the LSA process is that it often involves not only the verbatim transcription of the subject's spoken output, but also the corresponding representation using the International Phonetic Alphabet (IPA).

The IPA phonetic transcription is pivotal for conducting detailed phonetic-phonological analyses, which are essential when evaluating speech sound production, identifying patterns of phonological processes, and distinguishing between articulation errors and underlying linguistic deficits [40]. The use of IPA requires specific training and a high level of phonetic competence, further increasing the cognitive and operational demands placed on clinicians. Moreover, integrating phonetic data with broader linguistic measures within a single language sample analysis adds to the methodological and interpretative complexity, making the process time-consuming and potentially challenging to implement in routine clinical workflows. In response to these challenges, researchers and clinicians are increasingly exploring technological solutions that can streamline and enhance the assessment process. Advancements in technology are beginning to revolutionize speech therapy: Artificial Intelligence (AI) and Machine Learning (ML) offer new opportunities for earlier diagnosis and more personalized interventions, while teletherapy platforms and interactive applications have also expanded access to services, making therapy more accessible and engaging for young patients [46].

Among AI-driven innovations, Automatic Speech Recognition (ASR) systems are particularly promising for their potential to automate aspects of speech assessment. Current state-of-the-art models, such as OpenAI's Whisper [54], offer high levels of accuracy in converting speech into text, facilitating tasks like LSA and progress monitoring. However, most ASR systems are optimized for converting speech to standard orthography and are not designed to capture the phonetic detail necessary for clinical use in speech sound disorder assessment. A crucial and critical unmet need remains: the automated transcription of speech into IPA, especially in a way that preserves speech errors. Indeed, these errors are essential diagnostic features, as they convey a majority of clinically relevant information. Preserving them in transcription is therefore not optional but fundamental for capturing the linguistic patterns that inform clinical decisions. While some phonetic ASR models exist in the literature, they are largely focused on English and other well-resourced languages [60, 22]. To date, no open-source ASR model is specifically tailored for Italian, which presents a significant gap for clinicians working in that language. Accurate IPA transcription that reflects speech errors could provide SLPs with unbiased, fine-grained insights into children's articulatory patterns, supporting more precise diagnosis and the longitudinal monitoring of therapy outcomes. Bridging this technological gap is essential for extending the benefits of AI to a broader range of linguistic contexts.

It is important to note that ASR for child speech remains a challenging task and has attracted increasing attention from research community due to the substantial performance gap between recognizing adult and child speech: while modern ASR systems can approach human-level accuracy on adult voices in clean conditions, their accuracy degrades markedly on children's voices [5, 50]. In recent years, researchers have addressed this gap by identifying key obstacles in child speech transcription:

- Variability and Developmental Differences: Children's speech differs from adults' speech across acoustic, linguistic, and behavioral dimensions. Physically, children have shorter vocal tracts and smaller resonating cavities, leading to higher fundamental frequencies (pitch) and formant frequencies in their speech [51]. Recent analyses confirm that there are significant domain shifts both between child speakers and within each child's speech over time [55]. This intra- and inter-speaker variability is substantially higher for children, complicating the modeling task.
- Pronunciation and Linguistic Factors: Pronunciation differences are a prominent challenge. Children often have developing articulation abilities (they may substitute or skip certain sounds, speak with an interdental production of /s/ and /z/ etc), and generally exhibit higher pronunciation error rates than adults. Moreover, children's utterances can be less linguistically structured. They might use shorter phrases, unconventional grammar, or spontaneous interjections (e.g. hesitations, self-corrections, pauses, or generally take a longer time to pronounce the sentence they intend) that differ from adult speech corpora. The linguistic content also tends to differ: children's speech in interactive settings may include playful or imaginative language, incomplete sentences, and frequent code-switching [61].
- Acoustic Mismatch and Features: The acoustic profile of children's voices not only shifts mean pitch upward but also often results in broader spectral variance. Traditional speech feature extraction methods (e.g. MFCC) are derived from adult auditory models and may not optimally capture child speech characteristic [50]. Children's higher pitch and formant frequencies can lead to misalignment in feature spaces that ASR models trained on adult data expect. Moreover, the signal-to-noise ratio (SNR) might be lower for child speech if recordings occur in school or home environments with background noise or if children have less controlled microphone usage.
- Limited Training Data: A fundamental obstacle for child ASR is the scarcity of large, high-quality child speech datasets. Collecting and annotating children's speech is inherently difficult because it requires parental consent and ethical safeguards, and young children cannot record hours of speech as easily as adults. Additionally, privacy regulations around minors make sharing such data challenging. As a result, the publicly available child speech corpora are relatively small in size (often tens of hours, rarely a few hundred hours). For example, until recently, no open-source child speech dataset approached the scale of adult datasets like LibriSpeech (960 hours) or Switchboard. A survey noted that "a sizable open dataset for children's speech is still not available" in the broader research community [67], contributing to the lag in robust ASR for children. This data scarcity is a critical issue because state-of-the-art ASR models, especially end-to-end neural architectures, typically demand large training sets to generalize well. Most importantly, there is not a single dataset with native Italian child speech.

Addressing these challenges has become a central focus in recent research on child speech recognition. The present work originates from a collaboration between clinical experts and researchers in AI and natural language processing, aimed at developing a standardized assessment and transcription platform. Within this context, the Talkidz Project was designed to investigate language development in Italian children by analyzing linguistic samples from both typically developing individuals and those with diagnosed language disorders. The Talkidz Project focused on developing the Talkidz Platform, an innovative web-based platform designed to support clinical and research activities, where speech therapists can access the facilities to carry out the Talkidz Evaluation, a novel semi-structured elicitation test designed to collect spontaneous language samples from children through image description. This test aims at deriving various linguistic indexes, including phonological accuracy (e.g., percentage of correct words) or morpho-syntactic complexity (e.g., mean length of utterance). The Talkidz Evaluation includes 40 carefully designed illustrations, each selected to prompt speech production based on a combination of phonetic relevance, semantic frequency, and phonetic complexity, and is partially inspired by the work by Bortolini's work in "Tests for the Phonological Assessment of Infant Language" (PFLI) [11] (the experiment itself will be discussed in more detail in Section 3.2.1).

Phonetic-phonological analysis is fundamental for identifying rehabilitation goals, but the current process is entirely operator-dependent, requiring significant time for transcription and error classification: for this reason, the Talkidz Platform offers an AI-based phonetic transcription pipeline that eases the speech therapist from the task during validation. A key feature of such AI pipeline is *Broca*, a Conformer-based [62, 34], Deep Learning (DL) model, able to perform speech transcription while retaining the child's speech errors, as they are essential for detecting and understand the sounds they are able to elicit and those they lack due to possible speech impairments. *Broca* provides structured, machine-readable phonological information that integrates into longitudinal patient records and decision-support tools, reducing cognitive load for clinicians.

Beyond phonetic transcription, the Talkidz platform provides interactive dashboards featuring validated clinical metrics and statistics. These tools were developed with a co-design approach based on feedback from practicing Italian speech therapists and support evidence-based decision-making in both diagnosis and treatment planning. Talkidz thus fills a critical gap, offering a structured, validated framework that improves the quality and precision of speech therapy for Italian speakers.

### 1.1. Research objectives

Having outlined the principal limitation in the field of phonetic transcription, especially in child speech, this study focuses on addressing the gap in the literature concerning the accuracy of grapheme-based versus phoneme-based ASR models for the Italian language. This leads to the formulation of the first research question:

**RQ1:** To what extent is it feasible to develop an AI-driven transcription system that accurately transcribes Italian children's speech while preserving phonological and articulation errors?

In addition to addressing technical limitations in phonetic transcription systems for child speech, this study also investigates the potential of AI-driven transcription tools to support clinical workflows in speech therapy. Unlike traditional transcription by humans, which is subject to variability, automated systems offer the possibility of consistent and fast phonetic transcriptions. For these reasons, this research also aims to answer to the following question:

**RQ2:** How can AI-driven transcription ease speech therapist's workload, and in what ways can these insights enhance clinical practices in language assessment and rehabilitation?

In addition, this study investigates the role of dataset curation and phoneme selection in transcription performance:

**RQ3:** To what extent can a small but carefully curated, manually transcribed at the word level with a phoneme inventory dataset (restricted to the Italian language) outperform models trained on much larger but automatically transcribed corpora?

The whole project aims to make a first step toward a transformation of speech analysis in a more automated, AI-driven task. The following sections present a comprehensive overview of the research: Section 2 reviews related work from two perspectives: traditional approaches in speech therapy and recent advances in AI-based speech-to-IPA transcription tools, useful to frame the interdisciplinary nature of the project. Section 3 details the methodology used to develop and train the proposed model, including a description of the standardized speech assessment used to collect child speech data, while Section 4 reports the experimental results, evaluating model performance and comparing it to several state-of-the-art open-source models, including an interesing 2-steps approach using OpenAI's Whisper[1]. Finally, Section 5 discusses the conclusions drawn from the findings and provides answers to the research questions posed in the Introduction.

## 2. Related Works

This section provides an overview of recent technological advancements in two key areas: digital solutions for speech therapy and AI-based (ASR) for speech therapy. Notable progress includes the development of web platforms aimed at supporting therapeutic interventions, especially for children, alongside significant improvements in ASR systems for both adult and child speech. The following sections examine these developments in greater detail.

### 2.1. Speech Therapy Technologies

A systematic review by Deka et al. examined "AI-based automated speech therapy tools for persons with Speech Sound Disorders" [22]: part of the survey is dedicated to investigate the level of autonomy achieved by such models, and the analysis reveals that a significant number of studies have developed fully automated speech therapy systems. These systems operate often times independently, without incorporating the roles of SLPs, caregivers, or other stakeholders. For instance, the tool implemented by Ng et al. provide automatic feedback and assessments without human intervention [47], while Desolda et al. in [23] emphasize a collaborative approach, proposing a web-based platform named Pronuntia, allowing SLPs to assign personalized therapy exercises, caregivers to support practice at home, and children to engage interactively. This kind of approach is also present in the Talkidz Project, designed to keep the SLPs at the core of both diagnosis and therapy.

Deka et al. also posed the focus on speech therapy platforms, e.g. the tablet-based application designed by Ballard et al. in [6] to assist individuals with apraxia of speech and aphasia. The application utilizes ASR (at grapheme level) to provide real-time feedback on speech accuracy during therapy sessions. In their study, five participants engaged in a four-week at-home therapy program, practicing 100 trials per session, four times a week. The ASR system's accuracy was compared to human judgment, revealing an average agreement of approximately 80%.

Sztahó et al. developed a fully automated, computer-based system aimed at teaching speech prosody to children with hearing impairments [59]. The system provides visual feedback on aspects such as intensity (accent), intonation, and rhythm, facilitating the learning of prosodic features. By displaying visual cues corresponding to speech input, the tool aids children in understanding and producing appropriate prosodic patterns. The automated nature of the system allows for independent practice, making it a valuable resource for children requiring additional support in prosody acquisition. Participants demonstrated improved word production accuracy over time and reported positive experiences using the app alongside clinician support. Although the proposed platform, discussed in by Deka et al.[22], shows interesting features and promising results, the review highlighted a lack of studies comparing these tools to traditional speech therapy, calling for more research and clearer design guidelines.

The Talkidz Platform differs significantly from the approaches proposed in the aforementioned papers in both purpose and implementation, justifying the completely different perspective in designing and implementing AI-based tools: rather than serving as a tool for autonomous speech training or repeated practice exercises, our platform is designed to support the structured administration of a phonetic experiment under professional supervision. This guaranteess high-quality data collection and and allows the therapist to observe, interact, make immediate clinical judgments. Recent AI applications also include Deep Learning models for classifying dysarthria severity, as shown in Suresh et al.'s work with Deep Neural Networks (DNN) and Convolutional Neural Networks (CNN) [58].

## 2.2. Automatic Phonetic Transcription

In the recent years, advancements in ASR carried over the phonetic transcription task, a field where Transformer-based models [62] is ubiquitous.

Taguchi et al. [60] propose a universal ASR model that transcribes speech into IPA using a fine-tuning wav2vec 2.0 architecture [4], training on Common Voice [2] in many languages. Their system is shown to reach almost human-level performance in phonetic transcription, but both training and testing set are converted to IPA using Grapheme-to-Phoneme (G2P) tools. Since the transcription is heavily based on how well the G2P performs in such languages, the authors had to test their methods only on languages with strong mapping between graphemes and phonemes, thus underlying the crucial role of correct phoneme textual transcription. Furthermore, the study does not include Italian among the target languages, even though Italian is known for its relatively transparent orthography. This highlights the necessity for accurate phonetic transcriptions, particularly for languages like Italian, where such resources may be less developed. Although the present work shares certain similarities with the work from Taguchi et al., such as the training of a BERT-like model, it differs in key aspects, especially the use of G2P tools for training, which in this work is limited to the pretraining phase (see Section 3.1).

Yusuyin et al. [68] introduced a multilingual ASR framework that uses weakly supervised phonetic transcriptions. While being particularly effective in low-resource settings, outperforming subword and self-supervised learning approaches in crosslingual transfer tasks, the G2P transcriptions in the proposed framework are generated using Phonetisaurus [48], a finite-state transducer trained on LanguageNet data. Phonetisaurus has reported Phoneme Error Rates (PER) ranging from 7% to 45%, indicating potential inaccuracies in the phonetic transcriptions [36]. The present work differs from earlier approach because of the fine-tuning on child speech using expert-annotated phonetic transcriptions provided by speech therapists. The combination of weakly supervised pretraining and clinically curated fine-tuning enables *Broca* to achieve a better adaptation to the acoustic and phonetic variability of child speech.

While both relevant works cited hiterto can be compared to *Broca* for the fundamental approach to the task, neither of them is focused on child speech recognition.

### 2.2.1. Child Speech Recognition

A comprehensive evaluation made by Fan et Al. [24] in 2024 on speech foundation models for ASR in children recognized the unique challenges that children's waveform poses, such as higher pitch, limited data, error-prone dialogue and greater variabiliy. While their research is solely focused on English-based publicly available child speech datasets (such as MyST [52] and OGI Kids' Corpus [56]), similar challenges with data scarcity and the distinct acoustic-linguistic traits of child speech were also encountered in this work of research, justifying the necessity of in-house data acquisition and human evaluation of speech features prior to training end-to-end AI-based frameworks. Another relevant work made by Medin et Al. [10] investigates how WavLM [14], can be adapted for phoneme recognition in French child speech. This study demonstrates that fine-tuning self-supervised models on child-specific data can significantly improve recognition accuracy, highlighting the importance of adapting speech models to the unique characteristics of child speech (as adopted in this work). This approach led the authors to significant improvements, with transcriptions achieving PERs as low as 26.1%. While this work shares a methodological similarity with our research, their transcription strategy differs: automatic phoneme transcription is applied only to correctly pronounced words, whereas speech errors are manually annotated.

In contrast to [10], our research adopts a fully manual phoneme-level transcription approach, conducted by expert speech therapists, regardless of correctness, ensuring that all relevant speech patterns, including errors and accents, are accurately captured. Moreover, there are key differences in the study populations and the nature of the speech data: the ages of the participants involved in the work carried out by Medin et al. ranges from 5 to 8 years old, and engage in non-spontaneous reading tasks, while our study involves younger children (ages 3.5–6.5) participating in structured yet naturalistic speech assessment sessions. This distinction highlights our emphasis on capturing a broader range of developmental speech phenomena, which is a key in early intervention contexts and less constrained by task-specific vocabulary or reading proficiency.

By comparison, recent works in the likings of IPA-CHILDES & G2P+ [31] and ZIPA [69] represent advances in large, multilingual or cross-lingual phoneme recognition and phonemic data resources. IPA-CHILDES provides wide coverage of child-centered speech across 31 languages (including Italian), and its tools like G2P+ ensure consistent phonemic inventories via automatic orthographic-to-phonemic conversion. However, because transcription is automatic rather than manual, systematic errors made by the automatic transcriber are introduced, and nonstandard pronunciations or speech impairments are likely underrepresented or misaligned. In contrast, this work of research is distinct in focusing on manually annotated child speech, with a major focus on trascribing audio derived from speech therapy session, where errors, atypical articulations and incomplete productions will be faithfully transcribed during the creation of the dataset, rather than normalized or discarded.

ZIPA, likewise, leverages massive corpora (17,000+ hours), pseudo-labeling, and efficient architectures to push the state of phone recognition across many languages, but it does not explicitly include child speech or speech with impairments, and its transcriptions are not designed for detailed phonetic error capture or the fine-grained annotations that speech therapists need. Automatic G2P conversion, such as those used in IPA-CHILDES or ZIPA's large-scale resources, inevitably standardize input and obscure the kinds of deviations that are clinically meaningful, especially when children produce sounds outside the expected phoneme inventory or fail to complete words. More recently, models such as MultIPA [15], wav2vec2-XLSR-53 [4, 19] wav2vec2-LV-60 [65] when fine-tuned, have demonstrated state-of-the-art results in pronunciation assessment and phoneme recognition. MultIPA employs a multi-task setup to assess pronunciation quality, fluency, and prosody, excelling in learner speech scenarios where the goal is proximity to a canonical target. Likewise, wav2vec2-XLSR-53 (XLSR-53), when fine-tuned for phoneme recognition, leverages powerful multilingual self-supervised representations and achieves strong phonetic accuracy across several languages. When it comes to wav2vec2-LV-60 (LV60), the authors fine-tuned a multilingual wav2vec 2.0 model using phoneme mapping to transcribe unseen languages, outperforming prior zero-shot cross-lingual methods. Yet in each case, the focus is on canonicalization: automatically transcribed data and correctness-oriented scoring dominate, while error-retention is underemphasized, and moreover child speech is almost never discussed: neither framework addresses the problem of impaired child speech, where productions often involve incomplete words, nonstandard articulations, or sound substitutions that are not easily mapped to canonical phonemes. The methodology described in this work aims to substantially differentiate from the work hiterto discussed, explicitly retaining speech deviations by relying on expert speech therapists who annotate audio at phoneme level with word boundaries in order to retain full spectrum of either correct or disordered speech patterns. Moreover, the annotation phase further restricts the inventory to phonemes explicitly used in the Italian language [21], avoiding exotic diacritics.

## 3. Dataset

Central to this process is the acquisition, processing and transcription of a suitable dataset.

### 3.1. Pretraining Data (Adult Speech)

Since the limited data on child speech could keep *Broca* from learning the logic behind IPA transcription, the Multilingual Librispeech [53] (MLS, Italian Only) Dataset was utilized. The speech data was automatically transcribed using an in-house rule-based model specifically designed for phonetic conversion. To confirm the reliability of these transcriptions during the pretraining phase, we conducted a thorough validation process to detect and correct systematic errors manually. This step was essential to prevent the introduction of transcription bias, which could otherwise compromise the quality and generalizability of downstream learning. Although no publicly available tool currently offers a fully accurate and robust solution for transcribing Italian audio directly into IPA, and no G2P model exists that meets the linguistic precision required for Italian phonological representation, an in-house rule-based approach was selected as the most viable and option. This rule-based converter is based on simple conversions of famously known patterns that can be trascribed from grapheme to phoneme, leveraging Phonitalia [32] for known words and handling exceptions with fallback rules. This choice was motivated by the need to constrain the phonetic transcription to a predefined and simplified IPA subset, omitting language-specific diacritics and suprasegmental features, in order to maintain consistency across the corpus. This pragmatic compromise allowed us to generate phonetic transcriptions that, while not exhaustive in phonological detail, were well-suited for training and evaluating models under a controlled and linguistically interpretable phoneme inventory.

### 3.2. Fine-Tuning Data (Child Speech)

The entire child speech dataset used for fine-tuning was recorded during the execution of the novel Talkidz Evaluation, along with two additional standardized neuropsychological tests commonly used in speech therapy, namely the "Speech Evaluation Test for children aged 4 to 12" test by Marini et al. [44] and the "Neuropsychological Lexical Test" by Cossu [20], which are going to be discussed in detail in this section. While *Broca* is intended to transcribe speech specifically recorded during the Talkidz Evaluation, including recordings from the additional tasks helps reduce overfitting to the experiment’s limited vocabulary and promotes better generalization to spontaneous child speech. Recordings are conducted in acoustically controlled environments using high-fidelity microphones, which capture audio at 96.000 Hz, 24-bit, to minimize noise and distortion. Artificial noise can later be added to train models for real-world robustness. In this section, the details about the experiment itself will be discussed.

#### 3.2.1. Part I: The Talkidz Evaluation (Drawing Description)

The core of the Talkidz test is enclosed in this phase: 40 drawings have been proposed to elicit specific speech sounds in children. The test aligns with children’s cognitive and linguistic development, focusing on a controlled vocabulary and phonetic balance while avoiding rare or complex words. The selection of target words for our novel speech therapy test is based on semantic and frequency criteria, ensuring that the elicited words are both meaningful and accessible for children. To achieve this, critical words have drawn from the "Child’s First Vocabulary" (PVB) by Caselli et al. [13], a well-established lexical database for Italian-speaking children.

The words were specifically selected from the category of concrete nouns, chosen based on semantic criteria to facilitate their representation as images, making them easily recognizable and engaging for young participants. Furthermore, word frequency has been counted as a factor in the development of the test: only the most commonly used words in early vocabulary acquisition were included. To further ensure that every word could be elicited without introducing unfamiliar vocabulary, selected words have been empirically proven to be acquired within the first 30 months of life [13]. This approach reduces the likelihood of introducing unfamiliar or developmentally inappropriate vocabulary, which could lead to speech errors unrelated to underlying speech impairments.

The 40 illustrations used for language sample collection were designed following the criteria discussed by Zmarich et al. [70], namely:

1. Phonetic Criterion: The words were selected to elicit all Italian consonantal phonemes in at least two different test words for each positional context (word-initial singleton, word-medial singleton, word-initial cluster, and word-medial cluster).
2. Semantic/Frequential Criterion: Words are selected from the concrete noun category (semantic criterion) discussed in Appendix A of the PVB [13], to ensure they could be represented through images and were chosen based on the highest frequency values (frequential criterion). Additionally, to further ensure that all selected words would be familiar to the children, only those proven to be acquired within the first 30 months of life were included.
3. Phonetic Complexity Gradation Criterion: The selected words reflect a range of phonetic complexities, from simpler to more challenging syllable structures and consonant clusters.

The experiment is carried out as the following: during the test, children are asked to describe "what is happening" or "what do they see" in each of the 40 illustrations, and the only further prompt provided by the examiner, in case of non-exhaustive description by the child, is "Tell me something else". This prompt forced the child to express more thoughts on the illustration, in order to avoid closed-ended questions that might constrain the child's speech production to single words. The issue of how best to elicit spontaneous speech and the influence of materials or contexts used for language sample collection has been widely discussed in the literature, with no definitive consensus. One of the most widely recognized and validated studies on the topic distinguishes between maximum performance and typical performance [57]. The former represents the child's highest possible speech capability, which structured assessments aim to measure. According to the authors, story generation in a free-play context is the best way to elicit maximum performance. However, other studies have found no clear distinction between these two performance levels. For example, Mirsaleh et al. [66] reported similar results for the number of utterances produced across different elicitation contexts. The decision to use an image description task in this study reflects a compromise between the degree of control exerted by the examiner over the child's speech production and the child's actual verbal competence, which may vary in different communication contexts.

The administration of the illustrations took an average of 20 minutes per child (± 5.25 minutes). On average, each child produced 140 utterances (± 55 utterances).

#### 3.2.2. Part II: Speech Evaluation Test for children aged 4 to 12 (Denomination and Articulation Task)

The "Speech Evaluation Test for children aged 4 to 12" (BVL 4-12) by Marini et al. [44] is a clinical and diagnostic tool used to evaluate language skills in children aged 4 to 12. This battery is designed to assess different components of language, both in terms of production and comprehension, and can systematically assesses phonological, lexical, semantic, pragmatic, and discourse abilities through tasks involving production, comprehension, and oral repetition in children in the targeted age, with the aim of identifying possible communicative and linguistic disorders. Among the various subtests included in the battery, three key tasks for analyzing phonological and morphosyntactic abilities are used and recorded in order to both test the equivalent validity of Talkidz and to obtain more data:

**1.** Naming and Articulation

- This subtest assesses the individual's ability to correctly name presented images, testing lexical access, phonological production, and articulatory precision.
- It is particularly useful for identifying difficulties in lexical retrieval, word-finding issues (anomia), and articulation problems, which may indicate language or speech disorders.

**2.** Repetition of Nonwords

- This task is designed to examine the subject's phonological abilities by measuring their capacity to repeat meaningless words.
- Since nonwords lack inherent meaning, they minimize the influence of pre-existing lexical knowledge, allowing for a direct assessment of phonological memory and phoneme sequencing skills.
- This test is considered an important clinical marker of LD in both English and Italian [12].

**3.** Sentence Repetition

- This subtest evaluates the individual's ability to repeat sentences of varying syntactic complexity.
- It measures both short-term verbal memory and morphosyntactic and phonological skills.
- Difficulties in sentence repetition may indicate specific LD or linguistic deficits related to syntactic processing.

This battery is widely used in neuropsychological and speech therapy settings to assess and monitor language difficulties in different pathological contexts, including LD, verbal dyspraxia and ASD. Due to the variety of its subtests, the BVL 4-12 provides a detailed and multidimensional assessment of language, helping clinicians and researchers develop targe"..."ted rehabilitation and intervention programs.

#### 3.2.3. Part III: Neuropsychological Lexical Test (Semantic Fluency)

The "Neuropsychological Lexical Test" (TNL) by Cossu [20] is a semantic fluency assessment used to evaluate lexical retrieval ability and the semantic organization of vocabulary in individuals of various age groups, particularly in children. In the semantic fluency task, the child is asked to list as many words as possible belonging to a specific semantic category (e.g., "animals", "foods") within a set time limit, usually 60 seconds. The objective is to assess:

- The quantity of words retrieved (lexical productivity).
- The semantic organization of the lexicon, (i.e. how words are recalled and grouped).
- The efficiency of executive processes and lexical search.

This test evaluates various cognitive and linguistic abilities:

**1.** Access to the mental lexicon: the ability to retrieve words from one's vocabulary.

**2.** Semantic organization: the ability to group related words (e.g., naming terrestrial animals first and then aquatic ones).

**3.** Executive Functioning: the ability to maintain focus on the task, avoid repetitions and rule violations, such as producing words from an unrequested semantic category.

**4.** Cognitive Flexibility: the ability to quickly shift between different semantic subcategories.

#### 3.2.4. On Dataset Scale and Annotation Quality

Once data is collected, the next step is to manually transcribe the speech data into IPA notation. In this case Italian-specific IPA symbols are used, maintaining consistency with the phonemes generated by our in-house rule-based G2P converter deployed for transcription of the text derived from adult speech.

The phonetic transcriptions undergo a rigorous validation process, where each phoneme is checked manually by professional speech therapists with extensive experience in IPA transcription and speech therapy. In this setup it is imperative to keep a high phonetic accuracy: the model will strengthen the deep connections between audio and IPA characters learnt by transcribed samples from adult speech, and, in the presence of an error made by the in-house G2P model derived from an edge-case which mismatches the audio speech from the correct transcription, the model will be "rieducated" in order to fit the new, correct speech-to-text mapping. In order to keep a high transcription quality, any segment where phonemes are inaudible or speech cannot be reliably transcribed is removed: the objective is to eliminate every source of label noise in the training data.

A total of 21 children (10 females, 11 males, with an average age of 64 ± 8.6 months) participated in the data collection up to May 2025. From this dataset, a representative sample of 15 children, balanced by gender and age, was selected. Their speech was manually transcribed at the word level using the International Phonetic Alphabet (IPA), yielding 165 minutes of annotated audio for model fine-tuning[2]. To ensure consistency and reduce unnecessary complexity, the transcriptions were restricted to the set of phonemes explicitly used in Italian[21], avoiding exotic diacritics and nonstandard symbols. This focused inventory facilitates more reliable model training and better alignment with the phonetic targets relevant for Italian child speech.

While the overall duration can be perceived as modest compared to large-scale speech corpora, its size reflects several key constraints and deliberate design choices which are at the core of this work of research. First, the dataset prioritizes quality over quantity: every second of audio is manually transcribed by speech therapists to produce high-fidelity phonetic labels, avoiding unnecessary diacritics and ensuring compatibility with the limited Italian phonetic system. Importantly, the transcriptions are specifically performed at the word level, allowing the model to learn not only accurate phoneme generation but also correct word boundaries. This capability is rarely addressed in current state-of-the-art phoneme transcription systems (as discussed in Section 2.2), which typically output phoneme sequences without segmentation. For our application, however, word-level alignment is essential: the model is designed to be deployed in the Talkidz Platform, which provides automatic speech-to-IPA transcription for speech therapists, where both phonetic accuracy and word segmentation are necessary to compute the meaningful statistics about the Talkidz Evaluation. Second, data acquisition is inherently limited, as it involves children with speech impairments (a population that is not only small but also requires careful, ethically constrained recruitment and recording conditions). Third, the central goal of this work is to demonstrate that a carefully curated and precisely annotated dataset, though limited in scale, can yield comparable or even superior results to much larger corpora when those corpora rely on noisier, automatically generated transcriptions. Indeed, our experiments (discussed in Section 5) show that models trained on thousands of hours of automatically transcribed adult speech exhibit higher PER than *Broca*, even when their outputs are manually post-processed to normalize phoneme sets. This highlights the core contribution of our research: in the domain of child speech transcription, especially for speech impairments, high-quality phonetic annotation is more impactful than raw dataset size.

## 4. Methods

As outlined in the earlier section, the evaluation phase of our speech assessment protocol is fully delivered via the Talkidz Platform. Following this phase, however, speech therapists are tasked with manually transcribing each recorded utterance. This step typically requires the transcription of the child's speech in a way that preserves phonetic errors, a practice that can be tedious and difficult to scale in clinical or research settings. To address this, we aimed to develop an interface powered by *Broca*, tailored to the evaluation's specific needs. The proposed system allows therapists to upload speech recordings, which are then transcribed automatically by the model. The model will then automatically perform phonetic transcription, including the preservation of speech errors, with no need for further manual intervention.

This automation is intended to significantly streamline the analysis phase and allow therapists to focus more on interpretation rather than transcription.

The foundational architecture of *Broca* is W2v-BERT 2.0, a Conformer-based [34] speech encoder developed by Meta AI. This model is integrated into the SeamlessM4T v2 system [18], a multilingual and multimodal framework designed to support various speech and text translation tasks across more than 100 languages. Leveraging a unified architecture, it enhances processing efficiency and enables robust communication in diverse linguistic scenarios. W2v- BERT 2.0 consists of 24 Conformer layers and approximately 600 million parameters. It was pretrained on 4.5 million hours of unlabeled audio data covering 143 languages—including Italian—using self-supervised learning techniques.

The final layer of *Broca* is composed by a Fully Connected (FC) layer and is trained using Connectionist Temporal Classification loss [33]. To train *Broca*, a carefully designed training pipeline must be implemented, relying on a multi-stage approach to optimize performance on child speech data. Central to this process is the acquisition, processing and transcription of a suitable dataset, which ensures the model's ability to generalize across phoneme pronunciations. First, phonetic pretraining is performed on a large corpus of adult Italian speech automatically transcribed using a G2P tool. This is followed by fine-tuning on child speech data manually annotated by speech therapists. The training procedure and implementation details are then presented.

## 4.1. Model Training Procedure

Before training on the previously discussed dataset, a preprocessing step is applied to transform the raw audio waveforms into a format suitable for downstream processing by the Conformer. Specifically, prior to feeding the data into the model, the speech signals are converted into Mel-Spectrogram representations, with the number of Mel filterbank channels set to 80[3]. This transformation not only compresses the temporal and spectral features of the signal into a more tractable form, but also serves as a standard input representation in state-of-the-art ASR systems, including Conformer-based architectures such as *Broca*.

### 4.1.1. Pretraining Phase on Adult Speech

The model underwent an initial training phase on the Italian portion of the MLS dataset keeping every layer of the model unfrozen. This step was instrumental in introducing the Italian phonemes to the model and in training the language modeling head, which benefited from an explicit phonetic representation of speech data. Since the original MLS dataset was automatically transcribed, an additional mechanism was incorporated to detect potential anomalies in STT transcription.

Several ASR models for Italian transcription were employed (in the likings of Whisper and Wav2Vec 2.0) to verify dataset accuracy, and transcriptions significantly deviating from the original dataset were flagged for manual review. As a result, approximately three hour of audio were identified with incorrect transcriptions, which were manually listened to and corrected to ensure high-quality training data. The training process utilized a learning rate of $5e{-}5$ and a batch size of 64, with a linear decay scheduler and a warmup period of the learning rate comprising 10% of the total training steps.

The AdamW ($\beta1$ = 0.9, $\beta2$ = 0.99) optimizer was employed to handle weight updates efficiently, leveraging decoupled weight decay to prevent scale-dependent regularization [43]. Given that no significant overfitting was observed during training, dropout [37] was not applied, and the weight decay [39] was set to $1e{-}5$. The training dataset encompassed a total of eight days of audio data, while the validation dataset comprised sixteen hours.

#### 4.1.2. Fine-Tuning on Child Speech

Since the recordings originate from speech assessment sessions, they naturally contain both child speech and adult speech produced by the therapist. However, only the child speech is retained for model fine-tuning, while the adult speech is excluded from the dataset, to ensure that the evaluation truly measured the model's ability to generalize to child speech. This child-only selection process was also essential to prevent inflated performance metrics, as the model had already been trained on adult speech. The fine-tuning process obtained the best hyperparameter thanks to the hyperparameter tuning phase, and achieved the best results, which will be discussed in Section 5.

#### 4.1.3. Hyperparameter Tuning

The process of fine-tuning involved multiple training runs to explore the impact of different learning rates and frozen layer configurations on model performance. The hyperparameter search space included three different learning rates ($1e{-}5$, $5e{-}6$, $5e{-}7$) and variations in the number of frozen conformer layers (2, 4, 8, 10, 12, 14, 18, and 22), starting from the deepest layers. Each fine-tuning instance ran for 20 epochs to ensure sufficient adaptation and closely observed to avoiding overfitting. During experimentation, fine-tuning instances using the smallest learning rate (5e-7) failed to even marginally improve the results obtained with the pretrained model before fine-tuning it, indicating that this value was insufficient for the model to fit the dataset within the allocated epochs. The model struggled to update its parameters effectively, leading to stagnation in learning. This is clearly worth noting, particularly because the model is dealing with significant domain shifts (transitioning from adult to child speech).

Training instances with a learning rate of 5e-6 showed the best results, but while this learning rate produced interesting results in terms of PER, it was only effective when almost all layers were unfrozen. This suggests that a lower learning rate is beneficial when fine-tuning the entire model, allowing gradual adaptation without drastically altering existing representations.

However, given the fact that a possible desideratum would be to retain the adult speech performance obtained in the pretraining phase while adapting to child speech, a strategy that required unfreezing the whole model could be less favourable, as there could be a risk of degrading the already optimized adult speech accuracy. AdamW ($\beta 1$ = 0.9, $\beta 2$ = 0.99) remained the optimizer of choice.

Since overfitting was not observed in this stage, no weight decay or dropout was applied. The entire training process has been implemented leveraging Pytorch [49].

## 5. Results

This section presents the results obtained from both the initial training phase on adult speech and the subsequent fine-tuning on manually transcribed child speech. The evaluation focuses on the model's performance in phoneme recognition, comparing it against a state-of-the-art baseline and analyzing key metrics such as Phoneme Error Rate (PER) and its weighted variant, described below.

### 5.1. Metrics: The Phoneme Error Rate

In order to both evaluate *Broca* and compare it with readily available, open-source SOTA models, the main metric employed is the Phoneme Error Rate (PER). PER quantifies the difference between the predicted sequence of phonemes and the reference (ground truth) sequence by computing the minimum number of edits—substitutions $S$, deletions $D$, and insertions $I$—needed to transform one sequence into the other, and is a widely used metric in every major work on phoneme recognition [9, 26, 16, 17, 25]. This value is then normalized by the total number of phonemes in the reference sequence $N$. It is formally defined as:

$$PER = \frac{S + D + I}{N}$$

Furthermore, a variant of the PER is also going to be employed, in which phoneme substitutions are weighted based on their phonetic similarity, rather than treated as equally incorrect. In this approach, the cost of a substitution reflects the articulatory distance between the predicted and reference phonemes, so that errors involving highly dissimilar phonemes (e.g., /p/ →/a/) contribute more to the overall error than substitutions between similar phonemes (e.g., /t/→/d/). This is meant to provide a more linguistically informed measure of performance, particularly suited to child speech, where slight mispronunciations often involve acoustically or articulatorily related sounds. Given:

- $R$= $[r1, r2, \dots, rN]$ be the reference phoneme sequence (without spaces)
- $H$= $[h1, h2, \dots, hM]$ be the hypothesis phoneme sequence predicted by *Broca*

The mapping δ: R × H → [0, 1] is the phoneme similarity function, where:

$$\delta(r_i, h_i) = \begin{cases} 0 & \text{if } r_i = h_i \\ d(r_i, h_i) & \text{if } r_i \neq h_i \text{ (substitution } S\text{)} \\ 1 & \text{if } r_i \text{ is deleted or } h_i \text{ is inserted} \end{cases}$$

and the weighted Phoneme Error Rate ($wPER$) is computed as:

$$wPER = \frac{\sum_{(r_i, h_i) \in \text{align}(R,H)} \delta(r_i, h_i)}{N}.$$

The values of $\delta(\cdot, \cdot)$ are calculated using the articulatory feature-based distance functions provided by the PanPhon library [45], which quantifies differences between IPA segments using weighted or unweighted feature edit distances.

## 5.2. Model Evaluation and Comparisons

To evaluate the performance of the proposed model on child speech, experiments were conducted in two key configurations: (1) after pretraining on adult speech, to establish a baseline and assess the effectiveness of pretraining on mismatched data, and (2) after fine-tuning on a dedicated dataset of child speech, to measure domain adaptation capability. To contextualize our results, we compare *Broca* against several strong publicly available baselines. These include Whisper large-v2 [54], a widely used ASR model trained on over 600,000 hours of multilingual audio, W2V- LV-60-FT (LV60)[4], as well as W2V2P-XLSR-53-FT (XLSR-53)[5] and MultIPA[6]. The latter two models are recognized as state-of-the-art for phoneme-level transcription and are commonly used as benchmarks in IPA transcription research. The specific version of Whisper was chosen because it uses 80 Mel-Spectrogram channels, matching the audio processing configuration of *Broca*, while XLSR-53 and MultIPA process audio samples in raw waveform. Table 1 shows the differences in architecture and size among the considered models. Whisper's architecture includes 32 attention layers each in the encoder and decoder, whereas *Broca*, following a BERT-like design, consists of 24 encoder-only attention layers. The XLSR-53 model follows the Wav2Vec2.0 design, starting with a convolutional feature encoder that maps raw audio into latent representations, followed by a projection layer and a Transformer encoder stack with 24 self-attention layers. MultIPA adopts the same Wav2Vec2 backbone, also with 24 encoder-only attention layers, but differs in the final classification head, which outputs predictions over a smaller phoneme set. Similarly, LV60 shares the same encoder architecture with 24 Transformer layers, but employs a different output head aligned to its training vocabulary. It must be noted that while *Broca* shares the same Wav2Vec2.0 backbone as XLSR-53, MultIPA, and LV60, it nearly doubles the parameter count by replacing the standard Transformer encoder with Conformer layers.

**Table 1.** Architecture details and difference in size between models

| Model | Layers | Width | Heads | Parameters | Relative size |
|---|---|---|---|---|---|
| Broca | 24 | 1024 | 16 | 606M | 1.0x |
| Whisper (large-v2) | 32 | 1280 | 20 | 1541M | ~2.4x |
| w2v-XLSR-53 | 24 | 1024 | 16 | 316M | ~0.5x |
| MultIPA | 24 | 1024 | 16 | 316M | ~0.5x |
| w2v-LV60 | 24 | 1024 | 16 | 316M | ~0.5x |

It must be noted that Whisper performs speech recognition at the grapheme level. To enable comparison with our phoneme-based model, a rule-based G2P converter is applied to Whisper transcriptions, yielding a two-step pipeline. Alongside this setup, the previously described XLSR-53, MultIPA, and LV60 models are also evaluated. As the results demonstrate in Section 3.2.1, this two-step approach provides a more reliable strategy for transcribing impaired child speech in Italian, where data scarcity and labeling challenges make direct phoneme prediction less effective and less interesting with SOTA Speech-2-IPA models.

Table 2 presents a comparative evaluation of the three phoneme recognition models on a child speech test set: (1) Whisper large-v2 with rule-based grapheme-to-phoneme (G2P) conversion, (2) *Broca* pretrained on adult speech, and (3) *Broca* after fine-tuning on manually transcribed child speech. The performance is assessed using the two introduced metrics, PER and wPER, along with the individual contributions of insertion (I), deletion (D), and substitution (S) errors.

Whisper with G2P conversion yields the highest PER (33.24%) and wPER (27.52%), with notable deletion (17.69%) and substitution (10.32%) rates, indicating a significant loss of phonetic content and reduced recognition accuracy when applied to child speech. The pretrained *Broca* model improves overall performance, reducing PER to 27.76% and wPER to 23.08%, with markedly fewer substitution errors (7.44%) and insertions (2.58%), although deletions remain high (17.74%), likely reflecting its lack of exposure to child-specific phonetic patterns.

The fine-tuned *Broca* model achieves the best performance across all metrics, with a PER of 16.10% and a wPER of 13.36%. Deletion errors drop to 8.60% and substitution errors to 4.42%, suggesting the model's improved sensitivity to the phonological variability typical of children's speech. Interestingly, insertion errors show a slight increase (3.09%) compared to the pretrained version, possibly reflecting the model's attempt to capture previously unrecognized phonemic content. Overall, fine-tuning significantly enhances the model's capacity to produce accurate, phonemically faithful transcriptions of child speech.

In contrast, XLSR-53, LV60 and MultIPA, considered state-of-the-art in adult speech transcription, perform very poorly when evaluated on impaired child speech. Their PER values range from 49.72% (XLSR-53) to 57.37% (MultIPA), which correspondingly high wPERs (40 −43%). The most striking weakness is in deletion rates, which tops 31%. Such figures suggest that these models often fail to register child speech at all, discarding large portions of the signal rather than attempting to map them to plausible phonemes. This remarkable inability to account for child-specific phonetic patterns stands in sharp contrast to the performance of the fine-tuned version of *Broca*, which, despite being trained on less than three hours of audio, reduces deletion errors to 8.60% and outperforms large-scale baselines by a wide margin.

Furthermore, a closer investigation of substitution errors on transcriptions predicted by XLSR-53, LV60 and MultIPA yielded phonemes that are entirely alien to Italian phonology, often producing exotic or spurious symbols without linguistic plausibility. By contrast, *Broca* (when fine-tuned) produces substantially fewer substitutions, as it is designed to predict only within the Italian phonetic inventory.

These findings will be examined in greather depth in the qualitative analysis in Section 5.3, where examples are provided to illustrate the nature of such errors.

These results highlight the substantial improvements in phoneme recognition performance achieved through fine-tuning on carefully curated, in-house recorded child speech data. This targeted exposure substantially enhanced the model's ability not only to transcribe child speech more accurately (in the context of the Talkidz Evaluation) but also to perceive speech content that was previously unrecognized, as evidenced by the marked decrease in deletion errors. A comprehensive analysis of these findings is provided in the following section.

**Table 2.** PER and wPER obtained on the child speech test set with different models

| Model | PER | wPER | Insertions (I) | Deletions (D) | Substitutions (S) |
|---|---|---|---|---|---|
| Whisper (large-v2+G2P) | 33.24% | 27.52% | 5.23% | 17.69% | 10.32% |
| w2v-XLSR-53 | 49.72% | 40.72% | 1.29% | 30.49% | 17.95% |
| w2v-LV60 | 53.71% | 43.01% | 1.22% | 31.28% | 21.20% |
| MultIPA | 57.37% | 42.98% | 2.46% | 25.60% | 29.42% |
| Broca (pretrained) | 27.76% | 23.08% | 2.58% | 17.74% | 7.44% |
| Broca (fine-tuned) | 16.10% | 13.36% | 3.09% | 8.60% | 4.42% |

## 5.3. Qualitative Analysis of Transcription Accuracy

While quantitative analysis metrics such as PER and wPER provide a global view of model performance, they do not capture the clinical adequacy of phonetic transcriptions in the context of speech therapy. In particular, speech therapists require transcriptions that are both faithful to the child's production and constrained to the phonological inventory of the target language. To investigate whether current state-of-the-art models satisfy these criteria, we conducted a qualitative error analysis on transcriptions produced by XLSR-53, LV60, MultIPA and *Broca*.

Three representative 5-second child speech segments were selected from the test set. For each segment, the phonetic transcriptions from the four models were obtained. Speech therapists from the team then reviewed the outputs, comparing them to manually curated reference transcriptions. The experts annotated errors with respect to

- presence of phonemes outside the Italian phonetic inventory
- missing or collapsed word boundaries
- systematic misrecognitions affecti clinical interpretability.

**Table 3.** Qualitative comparison of transcriptions for three representative 5-second child utterances. Expert evaluations summarize the clinical utility of each model's output

| Model | Example transcriptions | Expert feedback |
|---|---|---|
| SLP transcription | 1. vedo una bimba ke salta sul:a pot:saNgera<br>2. pOi vedo uN ga-<br>3. uN gat:o ke a paura del:e api<br>4. kwa vedo una ragat:sa ke muota nel mare | - |
| XLSR-53 | 1. veltou na bimbo te cialte sula pocianghera<br>2. quele tonga un gato<br>3. N ěa něato ke paUda dele api<br>4. wa vEd una raěatsa ke mOta ne mala | The phonemes belong to the Italian IPA inventory; however, the letter-by-letter transcription slows down human validation and does not make clear which segment of the audio the transcription refers to. |
| LV60 | 1. ver una bimbA te Ùalte sula posandera<br>2. pe veto n va<br>3. un ě(g)ata ke pa:da dele pi<br>4. wa ved una: radZat mAta n ma:r | Letter-by-letter transcription with symbols often outside Italian IPA; false starts and fragments not preserved; output difficult to interpret for clinical purposes. |

| Model | Example transcriptions | Expert feedback |
|---|---|---|
| MultIPA | 1. pa una bimo kE lÙalta sula potCaNgEra<br>2. pE rto N ga<br>3. uN gat:o kE a paura dolE a:pi<br>4. una rakaţ:a ki wmota ni nima | IPA symbols correctly used, some allophones like /N/ captured; no word segmentation; output closer to plausible phonetic transcription but not directly usable for clinical evaluation. |
| Broca (fine-tuned) | 1. vedo una bimba ke salta sula pot:saNgera<br>2. pOi vedo uN ga<br>3. uN gat:o ke a paura dele api<br>4. kwa vedo una ragatsa ke mOta nel mare | Closely matches SLP reference; preserves word boundaries; all phonemes belong to Italian IPA; minor gemination errors; clinically usable with minimal post-editing. |

As reported in Table 3, XLSR-53 produces phonemes that largely belong to the Italian IPA inventory, and some sequences correctly capture the acoustic content. MultIPA and XLSR-53 output a continuous string of phonemes without any segmentation (e.g., "veltounabimbotecialtesulapocianghera"), while LV60 generates almost character-by-character sequences separated by spaces. For the sake of readability in Table 3, the words were manually re-segmented by the expert; this does not alter the numerical results reported in the previous quantitative evaluation.

Experts noted that, while the base phonetic recognition of XLSR-53 is acceptable when the speech is actually detected, the letter-by-letter or unsegmented phoneme representation complicates validation and reduces usability. Word-level segmentation is a highly desirable feature in this evaluation context because it provides not only phonemic information but also insight into the production of whole words, which is essential for analyzing consistent errors in child speech. LV60 yields transcriptions often including symbols not belonging to the Italian IPA and failing to preserve prosodic cues or false starts. This makes its output impractical for clinical use. MultIPA demonstrates better phonetic accuracy and correctly uses IPA symbols, capturing some allophones such as /N/, that can result tricky in the context of Italian speech transcription. Again, like XLSR-53, the lack of word segmentation limits clinical utility, as speech therapists (moreover in the context of the Talkidz Evaluation) require clear word boundaries to identify recurring patterns. In contrast, *Broca* closely matches the SLP reference. Word boundaries are preserved, all phonemes belong to the Italian IPA inventory, and minor gemination errors are systematically represented. Experts judged the output clinically usable with minimal post-editing, demonstrating that targeted training on expert-annotated child speech enables reliable transcription suitable for therapeutic purposes.

Overall, these observations confirm that while adult-oriented ASR models can recognize phonemes (altough with major deletions and recurrent substitution errors), they fail to provide the structured, word-level information necessary for speech therapy. Our model demonstrates that combining accurate phoneme prediction with correct word segmentation produces clinically actionable transcriptions for child speech evaluation.

### 5.4. Statistical Evaluation

To assess whether differences in PER among models were statistically significant, a non-parametric Friedman test [27] followed by pairwise Wilcoxon signed-rank tests [64] with Holm correction [38] has been conducted. These tests were selected due to the non-normal and paired nature of PER values obtained by inferencing on 5 seconds audio samples across the various models. The Friedman test tests the null hypothesis that such samples, coming from different sources (i.e. the models compared in 2) but associated to the same ground truth, have the same distribution. The Friedman test will be used to test for consistency among the predictions obtained by the different models also because of its robustness to homogeneitu of variance and violations of normality, and is defined as:

$$Q = \frac{12}{nk(k+1)} \sum_{j=1}^{k} \left( \sum_{i=1}^{n} R_{ij} \right)^2 - 3n(k+1) \quad (1)$$

where:

- n is the number of utterances,
- k is the number of models,
- Rij is the rank of model j on utterance i based on its PER.

Intuitively, the test sums squared ranks across models, adjusting for number of utterances and models. A significantly large Q rejects the null hypothesis that all models perform equally. The associated effect size for the Friedman test, Kendall's coefficient of concordance $W$, measure the degree of agreement among the rankings and is defined as:

$$W = \frac{Q}{n(k-1)}. \quad (2)$$

Values of W range from 0 (no agreement) to 1 (complete agreement). For pairwise comparieson, the Wilcoxon signed-rank test compares two models by analyzing the signed differences $d_i = x_i - y_i$ of PER per audio sample, excluding the values where $d_i = 0$. The differences are ranked by absolute value, assigning ranks $r_i$. The test statistic S is the sum of ranks corresponding to positive differences:

$$S = \sum_{i:d_i>0} r_i \quad (3)$$

The final statistic T is:

$$T = min(S, S') \quad (4)$$

where S′ is the sum of ranks for negative differences. For larger samples, T approximates a normal distribution with mean $\mu T = n(n + 1)/4$ and standard deviation $\sigma T = \sqrt{[n(n + 1)(2n + 1)/24]}$, allowing calculation of a Z-score:

$$Z = \frac{T - \mu_t}{\sigma_T}. \quad (5)$$

The Wilcoxon effect size r is then computed as:

$$r = \frac{|Z|}{\sqrt{N}} \quad (6)$$

where N is the number of samples. This allows quantifying the magnitude of the difference beyond statistical significance. Finally, Holm's sequential correction adjusts p-values to control the family-wise error rate in multiple testing. The Holm's correction has been selected over alternatives such as Bonferroni [3] or Benjamini-Hochberg (FDR) [7] based on its suitability for confirmatory, low-multiplicity testing. Holm's method controls the family-wise error rate (FWER) like Bonferroni but is uniformly more powerful, reducing the risk of false negatives without inflating false positives. Unlike FDR methods, which are intended for exploratory settings with many simultaneous hypotheses, Holm's sequential rejective procedure is more appropriate when performing a small number of post-hoc pairwise comparisons following a significant omnibus test. This will ensure rigorous statistical control while preserving sensitivity to meaningful performance differences between ASR models. The method orders raw p-values and compares each to $\alpha/(m - i + 1)$, where m is the number of hypotheses and i is the rank in the ordered list.

Applying the Friedman test across all six models (Whisper large-v2+G2P, XLSR-53, LV60, MultIPA, *Broca* pretrained, and *Broca* fine-tuned) yielded a test statistic of $Q$= 523.82 with $p$< 5.75 × 10−111 and Kendall's $W$= 0.126, indicating that the null hypothesis of equal performance can be decisively rejected. Although the effect size points to a moderate level of agreement among the rankings, the extremely low $p$-value confirms that the observed differences are systematic and not due to chance.

Post-hoc Wilcoxon signed-rank tests with Holm correction revealed that nearly all pairwise comparisons were statistically significant. Our fine-tuned model consistently outperformed every other system with complete separation of scores ($r$= ∞), underlying the decisive improvement. The pretrained variant of *Broca* also significantly outperformed Whisper ($p$= 0.0103, $r$= 0.219), although the effect size was small. By contrast, comparisons among XLSR-53, LV60, and MultIPA highlighted their relatively homogeneous and poor performance on the child speech task: LV60 and MultIPA did not differ significantly ($p$= 0.383, $r$= 0.063), while XLSR-53 showed only modest advantages over the two ($r$= 0.208–0.290).

These results demonstrate two pivotal points. First, state-of-the-art adult speech transcription models such as XLSR-53, LV60, and MultIPA not only underperform Whisper transcriptions with rule-based G2P conversion when evaluated on impaired child speech, but also cluster together at error rates that are far above those of *Broca*. Second, fine-tuning on less than three hours of child speech data yields dramatic improvements, completely separating the fine-tuned system from all other baselines and reducing errors by a wide margin.

A detailed breakdown of Wilcoxon signed-rank results is provided in Table 4, which reports test statistics, corrected $p$-values, and effect sizes for each pairwise comparison.

**Table 4.** Pairwise Wilcoxon signed-rank test results for PER for each possible pair of the models discussed (Whisper, XLSR-53, LV60, MultIPA, Broca-PT, and Broca-FT), with Holm correction and effect sizes (r)

| Comparison | Wilcoxon statistic (T) | Raw p | Holm-corrected p | Effect size (r) |
|---|---|---|---|---|
| Whisper vs Broca-PT | 6023.50 | $3.43 \times 10^{-3}$ | $1.03 \times 10^{-2}$ | 0.219 |
| Whisper vs Broca-FT | 1869.50 | $9.99 \times 10^{-21}$ | $8.99 \times 10^{-20}$ | ∞ |
| Whisper vs XLSR-53 | 3760.00 | $3.44 \times 10^{-13}$ | $1.72 \times 10^{-12}$ | 0.522 |
| Whisper vs LV60 | 3604.00 | $7.70 \times 10^{-14}$ | $5.39 \times 10^{-13}$ | 0.537 |
| Whisper vs MultIPA | 3637.50 | $1.07 \times 10^{-13}$ | $6.39 \times 10^{-13}$ | 0.534 |
| Broca-PT vs Broca-FT | 1791.50 | $4.23 \times 10^{-20}$ | $3.38 \times 10^{-19}$ | ∞ |
| Broca-PT vs XLSR-53 | 1656.50 | $5.78 \times 10^{-23}$ | $6.36 \times 10^{-22}$ | ∞ |
| Broca-PT vs LV60 | 1619.00 | $2.21 \times 10^{-23}$ | $2.65 \times 10^{-22}$ | ∞ |
| Broca-PT vs MultIPA | 1697.50 | $6.08 \times 10^{-23}$ | $6.36 \times 10^{-22}$ | ∞ |
| Broca-FT vs XLSR-53 | 193.00 | $1.23 \times 10^{-32}$ | $1.72 \times 10^{-31}$ | ∞ |
| Broca-FT vs LV60 | 52.00 | $1.44 \times 10^{-33}$ | $2.16 \times 10^{-32}$ | ∞ |
| Broca-FT vs MultIPA | 219.00 | $2.69 \times 10^{-32}$ | $3.50 \times 10^{-31}$ | ∞ |
| XLSR-53 vs LV60 | 6098.50 | $6.01 \times 10^{-5}$ | $2.40 \times 10^{-4}$ | 0.290 |
| XLSR-53 vs MultIPA | 7115.50 | $3.86 \times 10^{-3}$ | $1.03 \times 10^{-2}$ | 0.208 |
| LV60 vs MultIPA | 8683.00 | $3.83 \times 10^{-1}$ | $3.83 \times 10^{-1}$ | 0.063 |

### 5.5. Discussion

The nature of the comparison between results obtained from Whisper and *Broca* after pretraining on adult speech must be discussed thoroughly because it provides a framework for analyzing the nature of errors made by the two models. Since the G2P converter used for Whisper evaluation is the same as that used to generate phonemic targets during *Broca*'s pretraining, results obtained from both models should exhibit similar systematic characteristics in phoneme interpretation from graphemes. For instance, if both models produce the same phoneme-level error, and Whisper's intermediate grapheme transcription is correct, the error is likely attributable to the rule-based G2P mapping. In this case, the misalignment is not due to model misrecognition but to a deterministic mapping rule applied to both models. This is expected and, at least during the pretraining phase, tolerated: the purpose of using automatically converted labels is not to reach clinical-level accuracy, but to expose *Broca* to broad acoustic–phonemic correspondences, even though some specific mappings are not perfect. The systematic nature of these mappings will thus serve as a useful prior.

The key insight arises when *Broca*, after fine-tuning, is bound to diverge (with a stronger rate when feeding the model with more data in this phase) from such systematic errors, producing phoneme predictions that better match human-annotated child speech. This indicates the *Broca*'s capacity to refine its internal phonological representations and overcome inherited biases. In such cases, *Broca* demonstrates its ability to capture finer-grained phonetic patterns (even more so in presence of child speech), which are not accesible through grapheme-based recognition alone. Thus, the comparison not only validates the effectiveness of the direct speech-to-IPA approach but also highlights how fine-tuning on high quality data enables the model to correct for generalization limits introduced during pretraining.

When analyzing the results in Table 2, a significant observation emerges when analyzing the deletion rates. Whisper, despite its high-quality general-purpose transcription capabilities and its extensive training, fails to detect a large portion of the phonetic content in child speech. Its 17.69% deletion rate confirms the hypothesis that ASR systems trained exclusively on adult speech data struggle not only to transcribe children accurately, but even to recognize children's speech as speech. This effect is exacerbated in children with speech impairments, where phonemes may be variably distorted, leading to low acoustic confidence and dropped segments in the ASR output. The same phenomenon is present after pretraining *Broca*, which was trained on adult audio but paired with phonemes generated via our in-house G2P converter, with deletions reaching 17.74%. However, once *Broca* is fine-tuned on just 165 minutes of manually aligned child speech, deletion errors drop significantly, from 17.74% to 8.60%, showing a clear improvement in capturing child-specific phonetic patterns, including those with low articulation clarity: even limited exposure to child speech is sufficient to significantly improve recognition of challenging speech segments.

Substitution errors in *Broca* remain low across both pretraining and fine-tuning phases, indicating that when the model does recognize a phoneme, it generally predicts the correct one. In contrast, Whisper has a higher substitution rate (10.32%). Although one might initially suspect that these substitutions are largely caused by ambiguities in the rule-based grapheme-to-phoneme (G2P) conversion, for example, cases where a grapheme corresponds to multiple possible phonemes, the data might suggest otherwise.

Since wPER penalizes substitutions between phonetically similar phonemes less heavily on average, the high wPER achieved by Whisper indicates that most substitutions involve distant phonemes, i.e. phonemes that differ significantly in articulatory or acoustic terms. This suggests that the source of these errors does not lie in the G2P step, but in the initial transcription: Whisper appears to generate incorrect graphemic transcriptions.

Turning to the other state-of-the-art wav2vec-based models (XLSR-53, LV60, and MultIPA), the performance gap is even more striking. While these architectures are considered strong baselines in adult ASR, their phoneme error rates on child speech are markedly worse. All three models exhibit deletion rates above 25%, indicating that large portions of the input signal are effectively ignored or discarded. In practical terms, the models frequently fail to register children's productions as speech at all.

Qualitative inspection further revealed that substitution errors often produce phonemes not even part of the Italian IPA inventory, with confusional symbols appearing in transcriptions. In some cases, these correspond to "phantom" phonemes not justified by the acoustic evidence, underscoring the models' poor alignment with the target phonetic system. These results can be attributed to the way such models are trained: they rely on pseudo-labels generated through automatic G2P conversion from graphemes, without human verification and without exposure to prosodic aspects of speech.

As a result, systematic errors and biases in the G2P conversion are inherited during training, and the models lack the corrective supervision needed to converge toward clinically valid phonetic representations. Furthermore, the absence of reliable word boundaries in their output further undermines their clinical applicability. For speech therapists, word-level segmentation is essential: not only do phonemes carry information about articulatory accuracy, but the identification of consistent word-level patterns is critical for diagnosing and tracking speech impairments.

More broadly, adaptive and personalized intelligent systems have long been investigated as a means of tailoring information and services to individual users and contextual conditions [28]. In the present clinical context, this perspective motivates the use of structured phonetic evidence to support more personalized speech therapy interventions.

By contrast, *Broca*, once fine-tuned on a modest but expert-curated dataset, reduces deletion and substitution errors dramatically, preserves word boundaries, and restricts predictions to the set of valid Italian phonemes. The outcome is a transcription that is not only accurate numerically but also clinically interpretable. These results reinforce that direct modeling of Speech-to-IPA transcription can surpass traditional ASR + G2P pipelines, which, counterintuitively, can be seen as a more effective approach to using current state-of-the-art models for Italian child speech transcription.

## 6. Conclusions

This work explored the development of *Broca*, an AI-powered phoneme transcription system tailored for Italian child speech. It addressed key challenges in feasibility, accuracy, and the broader implications for phonological analysis and language assessment. As earlier defined, this work aimed to determine an answer to the following question:

**RQ1:** To what extent is it feasible to develop an AI-driven transcription system that accurately transcribes Italian children's speech while preserving phonological and articulation errors?

The findings demonstrate that an AI-based approach is feasible and effective, and the resulting model of such framework, *Broca*, demonstrated that accurate, automated transcription of children's speech is achievable, despite the acoustic variability typical of young speakers. Furthermore, while the system still relies on expert input for data annotation and model training, it greatly reduces manual transcription effort. Fine-tuning on even 165 minutes of child speech significantly improved performance, highlighting the promise of self-supervised learning, speaker adaptation, and efficient architectures for future scaling. Accurately capturing phonemes in child speech is complex, both due to natural pronunciation variability and limited available datasets but by focusing on a standardized, limited vocabulary and consistent labeling with Italian IPA, *Broca* achieved reliable phoneme prediction. We argue that the active involvement of speech therapists in both the phonetic annotation and the design of the system to reflect clinical priorities was the decisive factor in this performance. This collaboration shows that expert-curated, high-quality data can be more important than dataset size. It offers a model that can be replicated for data-efficient, human-centred AI in low-resource speech processing and clinical applications.

This brings the focus on the second research question:

**RQ2:** How can AI-driven transcription can ease speech therapist's workload, and in what ways can these insights enhance clinical practices in language assessment and rehabilitation?

*Broca* enables fast, structured analysis of phonetic patterns, supporting speech therapists in diagnosing errors and tracking progress over time. It reduces the workload of manual transcription and may support the development of more personalized speech therapy interventions. Beyond Italian, the approach is adaptable to other languages and low-resource settings, where phoneme-level ASR can support speech assessment, documentation, and early intervention, especially in underrepresented or multilingual populations. While certain phonological detail may be unrepresented due to constraints in the phoneme inventory, the model produces consistent and clinically interpretable outputs that support applications in both speech therapy and linguistic research. These findings provide a positive response to RQ2. While *Broca* shows strong performance after fine-tuning on a small child speech dataset, several limitations affect its scalability and real-world applicability. The dataset used—just 165 minutes from fewer than 15 participants—is small by DL standards. This brings the third research question:

**RQ3:** To what extent can a small but carefully curated, manually transcribed at the word level with a phoneme inventory dataset (restricted to the Italian language) outperform models trained on much larger but automatically transcribed corpora?

The results suggest that the scale of training data is not the decisive factor in clinical phonetic transcription. Rather, the linguistic adequacy of the phoneme set, the presence of reliable word-level annotations, and the domain-specific nature of the training data determine whether a model's predictions are both accurate and usable in practice. From a clinical perspective, even a relatively small investment in manual transcription yields gains that seem unattainable with any realistically achievable amount of noisy, automatically transcribed data, providing a strong response to RQ3.

Talkidz offers a structured, validated framework that improves the quality and precision of speech therapy for Italian speakers. This work emphasises the importance of a human-centered, co-design methodology, in which clinicians' expertise and feedback are essential for developing AI tools that are clinically meaningful and ethically aligned with real-world practice. Even though the results are showing a great deal of interesting findings, *Broca* does not come without its limitations. Although the two-phase training pipeline improved generalization, broader testing is needed to evaluate performance on more diverse and disordered speech. Expanding the dataset would reduce overfitting but is constrained by the manual effort required for expert transcription, currently performed by a limited team. Another limitation stems from the rule-based phonetic transcription of adult speech during pretraining. Such systems often miss linguistic exceptions, leading to inconsistent phoneme labels.

Future iterations could use *Broca*, fine-tuned on kids and on manually transcribed labels, for iterative relabeling or adopt data-driven alternatives, though this depends on broader phonetic coverage in training data. Computational demands also pose challenges: with 600M parameters, *Broca* requires significant resources for real-time inference, limiting use in clinical settings without dedicated infrastructure or cloud solutions.

Several key areas remain for future exploration and improvement of *Broca*. One of the most immediate priorities is data acquisition and expansion. The current dataset is limited to one hour of transcribed child speech from a small number of participants, which restricts the model's ability to generalize across different voices, accents, and speech patterns. Future work will focus on collecting and labeling larger datasets, encompassing a wider range of children, including those with diverse dialects, phonetic variations, and speech impairments.

Increasing the dataset size would not only improve model accuracy but also help mitigate overfitting to specific speech patterns, even though the participants being children presents further complexity in both data collection and engagement, while the transcription and annotation of their speech requires an extremely time-consuming effort due to the significant number of hours involved.

Also, enhancing the rule-based grapheme-to-phoneme converter used during adult speech pretraining: while this rule-based approach generally performs well, it fails to capture a small subset of linguistic exceptions and irregular phoneme mappings of the Italian language, because the latter has numerous phonological rules that make it difficult for a static rule-based system to handle all cases correctly. Refining this converter, or even replacing it with a data-driven approach, could improve the accuracy of the phoneme labels used during pretraining.

Future work could also explore the impact of multilingual training on phoneme transcription accuracy. Since languages with similar phonetic structures share many phonemes, training on multiple well-balanced datasets from languages closely related to Italian, such as Spanish, Portuguese, or French, could enhance model generalization. This could be particularly beneficial for low-resource child speech datasets, as a multilingual training approach might compensate for the scarcity of data by learning generalized phonetic representations that transfer well across languages.

Finally, an important next step is evaluating the model on a bigger subset of non-normative child speech, particularly in cases of moderate to severe speech impairments. The current study was conducted primarily on typically developing children, meaning that the model's ability to transcribe speech from children with atypical phonological development could be better evaluated with more data. Testing on more children with speech disorders will provide valuable insights into how well AI-driven transcription models can handle challenging speech patterns.

## Acknowledgments

This work was supported by Fondazione CRT (Fondazione Cassa di Risparmio di Torino) through the project *"Talkidz: un software per l'analisi del linguaggio"* (project reference 106465/2023.1740).

## Notes

1. OpenAI Whisper: https://openai.com/index/whisper
2. All recordings and linguistic data from child participants were collected within the framework of the Talkidz Project, following written informed consent provided by their parents or legal guardians. The consent procedure explicitly authorized the use of the child's voice and linguistic samples for scientific research and dissemination. The study received ethical approval from the Bio-Ethics Committee of University of Turin under the agreement number 0332723/2024.
3. For more information on the Mel-Spectrogram and its role in automatic speech recognition (ASR) pipelines, see [1].
4. https://huggingface.co/facebook/wav2vec2-lv-60-espeak-cv-ft
5. https://huggingface.co/facebook/wav2vec2-xlsr-53-espeak-cv-ft
6. https://huggingface.co/ctaguchi/wav2vec2-large-xlsr-japlmthufielta-ipa1000-ns

## Credit authorship contribution statement

Nicola Barbaro: Conceptualization, Methodology, Software, Data curation, Formal Analysis, Resources, Validation, Supervision, Writing - original draft, Writing - review & editing. Cristina Gena: Conceptualization, Methodology, Supervision, Writing - review & editing, Project Administration, Resources. Francesco Petriglia: Conceptualization, Data curation, Investigation, Validation, Resources. Andrea Meirone: Conceptualization, Resources, Project Administration. Alessandro Mazzei: Validation, Writing - review & editing. Arianna Viotti: Investigation.